# Polyharmonic Spline Packages: Composition, Efficient Procedures for Computation and Differentiation

Bakhvalov Y. N., Ph.D., Independent Researcher, Cherepovets, Russia

bahvalovj@gmail.com, ORCID: 0009-0002-5039-2367

**Abstract.**

In [1] it was shown that a machine learning regression problem can be solved within the framework of random function theory [12], with the optimal kernel analytically derived from symmetry and indifference principles and coinciding with a polyharmonic spline. However, a direct application of that solution is limited by O(N³) computational cost and by a breakdown of the original theoretical assumptions when the input space has excessive dimensionality. This paper proposes a cascade architecture built from "packages" of polyharmonic splines that simultaneously addresses scalability and is theoretically justified for problems with unknown intrinsic low dimensionality. Efficient matrix procedures are presented for forward computation and end-to-end differentiation through the cascade.

Reference [1] formulated the regression problem as an approximation task using the mathematical apparatus of random function theory [12]. It was shown that if one assumes (by the principle of indifference) that the probability measure on an infinite-dimensional function space possesses natural symmetries – translation, rotation, and scale invariance, together with Gaussianity – then the entire solution scheme, including the kernel form, the type of regularization, and the noise parameterization, follows analytically from these postulates.

We briefly recap the key elements of the solution in [1].

Let the training set be given by input vectors $x_1, x_2, \dots, x_k$ ($x_i \in R^n$) and output values $y_1, y_2, \dots, y_k$ ($y_i \in R$). The regression task is to find a function $f(x)$ such that

$$y_i = f(x_i) + u_i, \qquad (1)$$

where $u_1, u_2, \dots, u_k \in R$ are independent normal random variables with zero mean and variance $\sigma^2$ (chosen by the practitioner).

Viewing $f(x)$ as an unknown realization of a random function and assuming the probability measure on the infinite-dimensional function space has the symmetries discussed in [1], one obtains a solution of the form

$$f(x) = \sum_{i=1}^{k} \lambda_i k_f(x_i - x), \tag{2}$$

where $k_f(\tau)$ is

$$k_f(\tau) = \|\tau\|^2 (\ln(\|\tau\|) - b) + c, \tau \in R^n \tag{3}$$

and b and c may be treated as constants whose values can be estimated as approximations to the following functions:

$$b(\tau) = \frac{\sin(\omega_0 \tau)}{\omega_0 \tau} - \ln(\omega_0) - \gamma - \int_0^{\omega_0 \tau} \frac{\cos(\omega) - 1}{\omega} d\omega \tag{4}$$

where $\gamma$ is the Euler–Mascheroni constant,

$$c(\tau) = \frac{\cos(\omega_0 \tau)}{\omega_0^2} \tag{5}$$

Relative to the magnitudes of $\tau$ (pairwise distances between input vectors in the training set), the cutoff frequency $\omega_0$ – below which harmonics of the random function are absent – is chosen to be very small, so that the corresponding period

$$T_0 = \frac{2\pi}{\omega_0} \tag{6}$$

is, on the contrary, much larger than the typical interpoint distances in the inputs. For example, for $\|\tau\|$ in the range [0, 10] and $\omega_0 = 0.001$, [1] used b = 7.33054 and c = 1,000,000.

The coefficients $\lambda_i$ in (2) are determined by the linear system

$$\lambda = (K + \sigma^2 E)^{-1} Y \tag{7}$$

where $K$ is the $(k \times k)$ matrix with entries $k_{ij} = k_f(x_i - x_j)$, $i,j = \overline{1,k}$

$E$ is the identity matrix;

$\lambda$ is the column vector $(\lambda_1, \lambda_2, \dots, \lambda_k)$;

$Y$ is the column vector $(y_1, y_2, \dots, y_k)$ from the training set;

$\sigma^2$ is the variance of the random variables $u_i$ in (1).

In practice, b and c can be estimated once; the main working formulas are (1), (2), (3), and (7).

The parameter $\sigma^2 > 0$ in (7) guarantees invertibility. For exact interpolation ($\sigma^2 = 0$), even under unfavorable point configurations one can use the exact definitions (4) - (5) in place of treating b and c as constants in (3).

The function $k_f(\tau)$ is a generalized covariance for a random function from the class of Intrinsic Random Functions of order 1 (IRF(1), Matheron [10]), of which $f(x)$ is one realization. At the same time, the term $\tau^2 ln(\tau)$ and its linear combinations are known as the polyharmonic spline, specifically the thin-plate spline [4], [9].

Note that $k_f(0)$ is also the variance of the random function. Simultaneously multiplying $k_f(\tau)$ in (3) and $\sigma^2$ in (7) by the same nonzero factor yields the same solution $f(x)$ in (2).

The formulation in [1] considered the scalar-output case $y_i \in R$, i.e., a single function $f(x)$. Now consider vector-valued outputs $y_i \in R^m$. To solve the approximation problem one must compute m functions $f_1(x), f_2(x), \ldots, f_m(x)$. These functions can be computed independently using (1) - (7). However, even if their values are unrelated, there is a computational linkage: all of them have the form (2) and differ only in the coefficients $\lambda_i$. In (7), the matrix $K$ is the same in all cases, and if one adopts the same variance $\sigma^2$, then the inverse $(K + \sigma^2 E)^{-1}$ is shared as well.

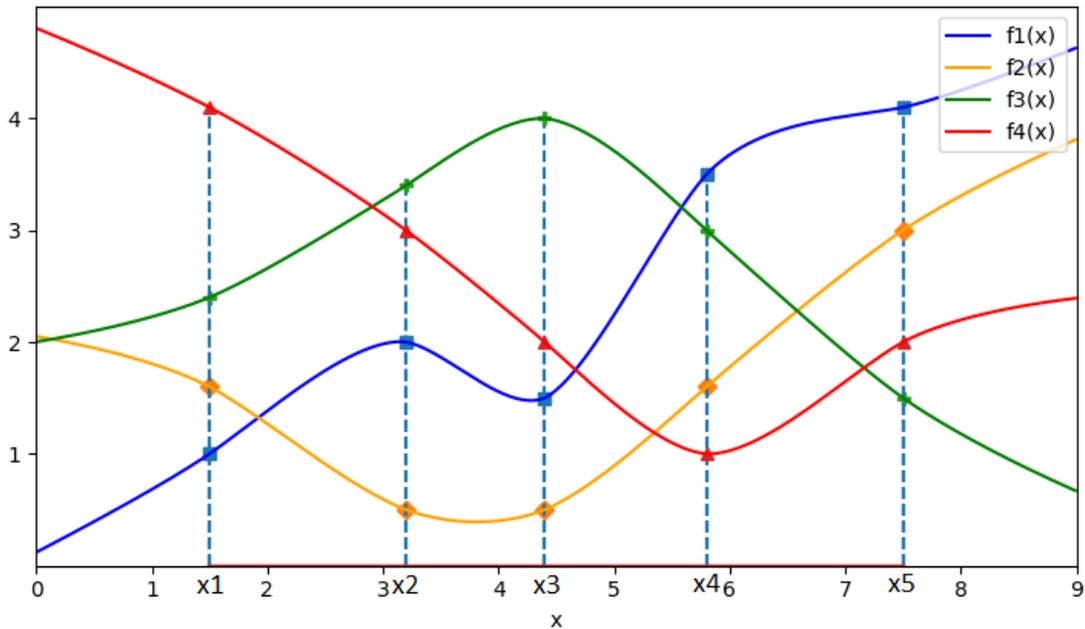

Figure 1. An illustrative one-dimensional example.

Here we present an example of constructing four different functions (the interpolation case with $\sigma^2 = 0$). They differ substantially from one another, yet all are built on the same set of input coordinates $x_1, x_2, \ldots, x_5$. All four functions are described by the same expression (2); they differ only in the coefficients $\lambda_i$, which in turn are computed by (7) using the same inverse matrix.

Thus, if we fix a set of key points $c_1, c_2, \ldots, c_k$ (we will call them a "constellation") and compute once the matrix

$$U = \left(K^{(c)} + \sigma^2 E\right)^{-1}, \tag{8}$$

where $K^{(c)}$ has entries $k_{ij}^{(c)} = k_f(c_i - c_j)$ evaluated via (3),

then within this constellation we can define any number of functions, each of the form (2), but parameterized via the constellation points.

Let the number of functions be m. Then we can write

$$f_t(x) = \sum_{p=1}^{k} \lambda_{pt} k_f(c_p - x), t = \overline{1, m} \tag{9}$$

Specifying any one of these functions amounts to prescribing its values (as a column vector) at the constellation's key points (or approximate values if $\sigma^2 > 0$) and multiplying that vector by $U$ from (8) to obtain the coefficient vector $\lambda_t(\lambda_{1t}, \lambda_{2t}, \ldots, \lambda_{kt})$.

In general, for $m$ functions this yields the $k \times m$ coefficient matrix $\Lambda$:

$$\Lambda = UY^*, \tag{10}$$

where $Y^*$ is the $k \times m$ matrix $[y_1^*, y_2^*, \ldots, y_m^*]$, and each $y_t^*$ is the column vector of prescribed values for $f_t(x)$ at the constellation points; $\Lambda$ collects the coefficients $\lambda_{pt}$ from (9).

Equation (9) can be written in matrix form as

$$F_x = K_x \Lambda, \tag{11}$$

where $K_x$ is the row vector $[k_f(c_1 - x), k_f(c_2 - x), \ldots, k_f(c_k - x)]$,

$F_x$ is the row vector of outputs $[f_1(x), f_2(x), \ldots, f_m(x)]$.

In what follows, we call the set of functions defined in this way over a given constellation a package. That is, by a package of polyharmonic splines we mean a family of functions of the form (2) built on a common set of centers, whose coefficients are determined via (10) using the shared inverse matrix (8).

Let us now detail how to work with this object computationally. Suppose we solve an approximation problem in $R^n$ using a package of $m$ functions

$f_1(x), f_2(x), \ldots, f_m(x)$, with $x \in R^n$. Represent the constellation as a matrix $C$ whose rows $c_1, c_2, \ldots, c_k \in R^n$ are the constellation points.

Rewrite (3) in a form convenient for computation:

$$k_f(\tau) = \tau^2(\ln(\|\tau\|) - b) + c = \tau^2 \left(\frac{1}{2}\ln(\tau^2) - b\right) + c =$$

$$= \frac{1}{2}\|\tau\|^2(\ln(\|\tau\|^2) - 2b) + c \qquad (12)$$

Note. The expression $\|\tau\|^2 \ln(\|\tau\|)$ is undefined at $\|\tau\| = 0$. However, since $\lim_{\|\tau\| \to 0} \|\tau\|^2 \ln(\|\tau\|) = 0$, one should explicitly set $k_f(0) = c$. For very small separations $\|\tau\|$ (e.g., $\ll 10^{-10}$) it is also advisable to use the approximation $k_f(\tau) \approx c$ to avoid numerical instabilities.

Expression (12) is convenient because $\tau$ enters only through $\|\tau\|^2$. To assemble $K^{(c)}$ in (8), first compute the matrix of squared pairwise distances between constellation points (denote it $M_c \in R^{k \times k}$). The law of cosines in matrix form gives

$$M_c = N_c J_{1,k} + J_{k,1} N_c^T - 2CC^T, \qquad (13)$$

where $N_c$ is the column vector of squared norms of the constellation points:

$$N_c = (C \circ C) J_{n,1}, \qquad (14)$$

∘ denotes the Hadamard (elementwise) product;

$J_{1,k}$ is the 1×k row vector of ones;

$J_{k,1}$ is the k×1 column vector of ones;

$J_{n,1}$ is the n×1 column vector of ones.

We can now compute the entries of $K^{(c)}$:

$$k_{ij}^{(c)} = \frac{1}{2} m_{ij}^{(c)} \left(\ln\left(m_{ij}^{(c)}\right) - 2b\right) + c, \qquad (15)$$

where $k_{ij}^{(c)}$ is an entry of $K^{(c)}$ from (8), and $m_{ij}^{(c)}$ is the corresponding entry of $M_c$ from (13).

With $K^{(c)}$ known, we compute U from (8) and then, via (10), obtain in one shot the coefficients (matrix $\Lambda$) for any desired number m of functions in the package.

Let us now consider the case where we need to evaluate a package of functions not at a single value x (as in (2) or (9), (11)), but simultaneously at many inputs. This is useful when processing data in mini-batches.

Suppose the package receives $r$ input vectors $x_1, x_2, \ldots, x_r$ ($x_i \in R^n$), collected into a matrix $X \in R^{r \times n}$, with each row being one $x_i$. For each of these, we need to compute the package's output vectors $y_1, y_2, \ldots, y_r$ ($y_i \in R^m$), each of dimension $m$, equal to the number of functions in the package. Collect these vectors into the output matrix $Y \in R^{r \times m}$.

Analogously to (13), first compute the matrix of squared distances between all rows of $X$ and all constellation points in $C$:

$$M_{xc} = N_x J_{1,k} + J_{r,1} N_c^T - 2XC^T, \qquad (16)$$

where

$$N_x = (X \circ X) J_{n,1}, \qquad (17)$$

$$N_c = (C \circ C) J_{n,1} \text{ (as in (14))}.$$

Unlike (13), $M_{xc}$ is rectangular, of size $r \times k$. Next, in analogy with (15), form the rectangular matrix $K^{(xc)}$:

$$k_{ip}^{(xc)} = \tfrac{1}{2} m_{ip}^{(xc)} \left( \ln \left( m_{ip}^{(xc)} \right) - 2b \right) + c, \qquad (18)$$

where $k_{ip}^{(xc)}$ is an entry of $K^{(xc)}$,

$m_{ip}^{(xc)}$ is the corresponding entry of $M_{xc}$ from (16),

$$i = \overline{1, r}; \; p = \overline{1, k}.$$

We can now compute the outputs:

$$Y = K^{(xc)} \Lambda \qquad (19)$$

Thus we obtain a package of polyharmonic splines of a special form – a universal construction for evaluating an arbitrary number of multivariate functions – where, to specify each function within the package, it suffices to prescribe its values at the constellation's key points.

It is natural to connect several packages of this type sequentially into a cascade, analogous to stacking layers in multilayer neural networks (Figure 2). In contrast to standard multilayer perceptrons, where nonlinearity is introduced via activation functions, the proposed cascade preserves global smoothness at every level by virtue of polyharmonic kernels. This makes it closer in spirit to deep kernel machines (Cho & Saul, 2009 [5]) or hierarchical RBF networks (Fasshauer, 2007 [7]), with a key distinction: here the basis functions are not chosen empirically but derived from symmetry principles, as shown in [1].

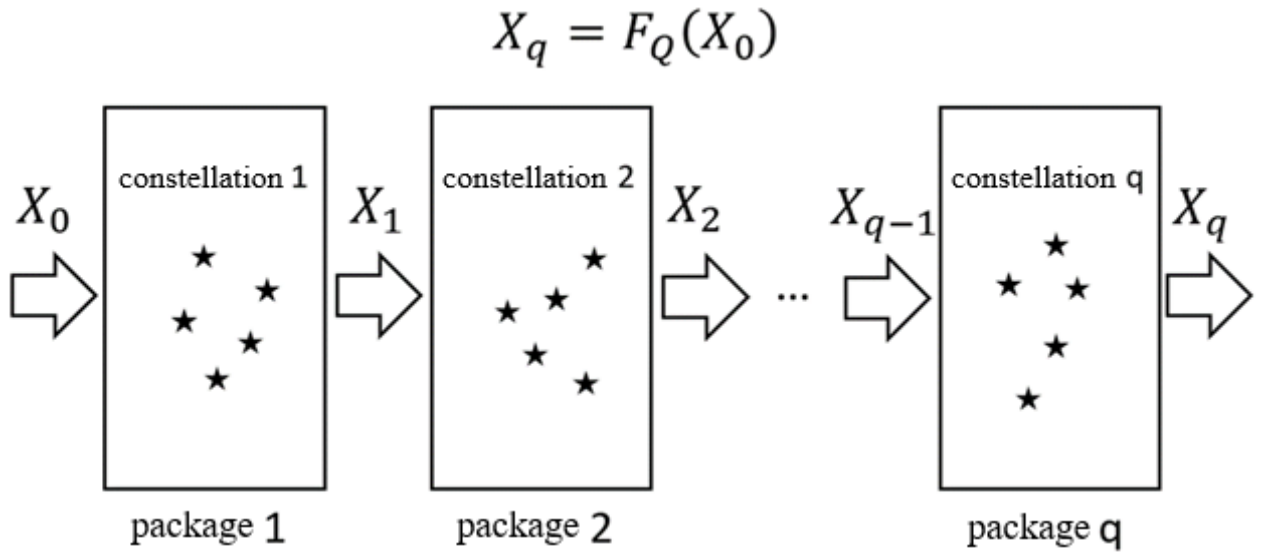

Figure 2.

However, a question arises. In [1] it was shown that expressions (1) - (7) provide a solution over the space of all continuous functions. Consequently, even a single package imposes no restrictions on approximating multivariate nonlinear functions of arbitrarily high complexity. Moreover, under the conditions described in [1], the solution is guaranteed to be optimal in a certain sense.

Why, then, build a sequence of packages? There are several compelling reasons; we list the most important ones.

The first reason is that in (2) (or in (9)) the number of terms equals the size of the training set. The same role is played by the number of vertices in the constellation when working with a package. Accordingly, the number of rows and columns in the inverse matrix computed in (7) - (8) grows with this size; the number of elements (and thus memory footprint) grows quadratically, while the computational cost grows cubically (or thereabouts).

Therefore, once the training set exceeds on the order of 10–100 thousand examples, solving the regression problem directly by simply increasing the number of terms in (2) becomes infeasible. Likewise, relying on a single package of polyharmonic splines and indefinitely increasing the number of constellation points (instead of cascading several packages) appears to be a dead end.

But the issue is not only computational. For many practical regression problems in machine learning, a sequence of packages can deliver better accuracy than a single layer. We now explain why.

In many real-world tasks, the number of features (the dimensionality of the input vectors $x_i$) is in fact redundant. Some features may be uninformative and unrelated to the target outputs; others may be mutually correlated and carry the same information. Thus, there often exists a mapping that transforms the original parameter space into another space of much lower dimension in which the approximation problem can be solved without loss of quality. This transformation may be unknown; for our purposes, it is the very fact of its existence that matters.

Let us consider what this implies when we return to the random-function-based solution that led to (1) - (7). First, we restate the three symmetry properties of the probability measure $\mu$ on the function space that underpinned the derivations in [1].

1. Invariance of the probability measure under translations and rotations of the coordinate system.

For any rotation matrix $M$, any translation vector $t$, and the mapping $T: f_1 \to f_2$ defined by

$$f_2(x) = f_1(Mx + t), x, t \in R^n \tag{20}$$

and for any measurable subset $A \subset \mathcal{F}$, one must have $\mu(A) = \mu(T(A))$. That is, any two subsets of functions that are mapped into one another by a rotation or a translation must have the same probability under $\mu$.

2. Invariance of the probability measure with respect to scaling transformations.

For any $k \in R$ and the mapping $T: f_1 \to f_2$ defined by

$$f_2(x) = kf_1(x/k), x \in R^n \tag{21}$$

and for any measurable subset $A \subset \mathcal{F}$, one must have $\mu(A) = \mu(T(A))$. In other words, any two subsets of functions related by a change of scale must carry the same measure $\mu$.

3. The probability measure μ is an infinite-dimensional Gaussian measure with zero mean.

This means that any finite-dimensional projection of the random function is multivariate normal, and the measure is Kolmogorov-consistent (Gikhman & Skorokhod [8]).

To better clarify item 3, [1] formulated two corollaries with an intuitive interpretation.

Corollary 3.1. From the Gaussianity of the probability measure μ it follows that if a subset of functions is generated by scaling a single base function $f_1(x) \to kf_1(x)$, then the conditional probability density on this subset is Gaussian in the parameter k with zero mean. In other words, if we assume the solution lies within the family

$$f_k(x) = kf_1(x), k \in R \tag{22}$$

where $k$ is a real scalar (we may call it the amplitude of $f_1(x)$) and $f_1(x)$ is any fixed continuous function used to generate the family (22), then the conditional distribution over this subset is normal in the amplitude k with mean zero.

Corollary 3.2. If we add an arbitrary fixed function $f_2(x)$ to each function in the set of Cor. 3.1, the distribution over $k$ remains Gaussian (although the mean may shift). Adding the same $f_2(x)$ to all functions yields a new family

$$f_k^*(x) = kf_1(x) + f_2(x), k \in R \qquad (23)$$

where $k$ is a real number (the amplitude of $f_1(x)$), and $f_1(x)$, $f_2(x)$ are any continuous functions.

Whatever $f_2(x)$ we choose, the probability density on this subset (as a finite-dimensional projection induced by $\mu$) remains normal in $k$, although the mean need not be zero.

Now consider a situation in which we a priori expect that some input features are redundant (or uninformative). As an example, take a function of two variables that we intend to approximate using the algorithm based on (20) - (23) (thus we may regard it as a realization of a random function):

$$y = f(x_1, x_2)$$

Suppose we know with very high probability (though less than 1) that one of the variables, either x₁ or x₂, is uninformative. That is, $y = f(x_1, x_2)$ is in fact either $y = f(x_1)$ or $y = f(x_2)$, although we do not know which. We denote these as $f(x_1, \_)$ and $f(\_, x_2)$, meaning the function remains two-dimensional but, in the first case, does not depend on $x_2$, and in the second case does not depend on $x_1$. Only with very small (but nonzero) probability do we expect $y = f(x_1, x_2)$ to be a genuinely nonlinear function of two variables.

Viewing $y = f(x_1, x_2)$ as an unknown realization of a random function, the candidate solutions can be any continuous bivariate functions, among which the families $y = f(x_1, \_)$ and $y = f(\_, x_2)$ appear as subsets. It follows that the probability measure should assign substantially higher mass to these subsets compared to other candidate solutions.

Take two nonlinear functions from these subsets, $f_1(x_1, \_)$ and $f_2(\_, x_2)$. Introduce their difference:

$$f_3(x_1, x_2) = f_2(\_, x_2) - f_1(x_1, \_) \qquad (24)$$

Clearly, $f_3(x_1, x_2)$ is a nonlinear function of two variables, and its value depends on both $x_1$ and $x_2$.

Now consider the subset of functions defined by

$$f_k(x_1, x_2) = f_1(x_1, \_) + k f_3(x_1, x_2), \tag{25}$$

where $k$ is a real number.

The family (25) specifies a subset for which the finite-dimensional projection of the measure applies and thus falls squarely under Cor. 3.2 (23). Substituting (24) into (25) gives

$$f_k(x_1, x_2) = f_1(x_1, \_) + k\big(f_2(\_, x_2) - f_1(x_1, \_)\big) \tag{26}$$

Then, for k = 0,

$$f_k(x_1, x_2) = f_1(x_1, \_), \tag{27}$$

and for k = 1,

$$f_k(x_1, x_2) = f_2(\_, x_2) \tag{28}$$

Thus, at $k = 0$ the function $f_k(x_1, x_2)$ depends only on $x_1$, and at $k = 1$ only on $x_2$. For other values of $k$ in this subset, the dependence is nonlinear in both variables. Under our assumptions, we would therefore expect the probability density over this subset, as a function of $k$, to exhibit two local maxima at $k = 0$ and $k = 1$. This contradicts Cor. 3.2 (23).

Therefore, we can no longer treat the random function as having a Gaussian measure on the function space. In this case, one cannot rely on the canonical and spectral representations of the random function and the subsequent derivations. Hence, the solution given by (1) - (7) can no longer be regarded as the natural choice, under the indifference principle, for approximating such a function.

We now sketch qualitative considerations for the probability measure in this setting.

In [1], the distribution over realizations of the random function in the function space was taken to be a Gaussian measure. In that case, the level sets of any finite-dimensional projection are hyperellipsoids whose principal axes align with the functions $e^{iwx}$ (if one uses the complex spectral representation).

By contrast, when we a priori expect the input feature space (in which the regression is posed) to be redundant, and when it can be relatively easily transformed (e.g., by discarding uninformative features or via transformations close to linear) into a space of much lower dimension in which the same regression task can be solved without loss, then hyperellipsoids are no longer an appropriate shape for the level sets of the finite-dimensional projections of the probability measure.

Functions that depend on fewer features (and functions that can be transformed into such by some multidimensional rotation) should carry higher probability density and form "protrusions" on the level sets compared to higher-dimensional functions.

In this case, the level sets resemble a multidimensional "hedgehog" or "octopus" rather than hyperellipsoids. In other words, the density becomes multimodal with peaks concentrated on lower-dimensional subspaces.

A rigorous construction of such measures lies beyond the scope of this paper; here we limit ourselves to proposing a computational architecture motivated by these considerations. Although these are qualitative arguments, they suggest that when the original feature space is redundant, the solution from [1] cannot be strictly regarded as the preferred choice under the absence of prior information.

However, if we approach approximation via a cascade of packages of polyharmonic splines, the picture changes.

Consider an n-dimensional space with a random function possessing the symmetries (20) - (23). Suppose we subsequently observe $m$ ($m \ll n$) realizations of this random function. On the set of values of these functions (whose dimension is less than $n$), we can again define a random function with the same symmetries (20) - (23). This procedure can be iterated multiple times.

The entire cascade can then be viewed as a single random function. How, then, will probabilities be distributed among the realizations of such a composite random function? Will the symmetry properties (20) - (23) of the probability measure on the function space still hold for this composite?

Let us denote the composite random function by $F_Q(x)$, comprising q layers of random functions (as above), $F_1(x), F_2(x), \ldots F_q(x)$:

$$F_Q(x) = F_q\left(F_{q-1}\left(\ldots \left(F_1(x)\right)\right)\right) \qquad (29)$$

Since $F_1(x)$ is spatially invariant – independent of the choice of origin and orientation – the composite function (29) is spatially invariant as well. Thus, item 1 corresponding to (20) holds for the composite random function. Corollary 3.1 (22) also carries over: the probability distribution on any subset of realizations that differ only by amplitude remains Gaussian in that amplitude (with zero mean). For this, it suffices that Cor. 3.1 holds for the random functions within $F_q(x)$.

What about the second property (21)?

Consider a particular realization $f_Q(x)$ of the composite random function (29):

$$f_Q(x) = f_q\left(f_{q-1}\left(\ldots \left(f_1(x)\right)\right)\right), \qquad (30)$$

where $f_1(x), f_2(x), \ldots f_q(x)$ are specific realizations taken by the random functions $F_1(x), F_2(x), \ldots F_q(x)$.

Strictly speaking, by $f_i(x)$ we mean a collection with as many components as the number of outputs of layer i.

The probability of observing $f_Q(x)$ as this specific combination in (30) equals the probability of simultaneously observing the corresponding realizations $f_1(x), f_2(x), \ldots f_q(x)$ that compose it.

Now consider the realization $f_Q^*(x)$:

$$f_Q^*(x) = f_q^* \left( f_{q-1}^* \left( \ldots \left( f_1^*(x) \right) \right) \right), \qquad (31)$$

where

$$f_i^*(x) = k f_i \left( \frac{x}{k} \right), \qquad (32)$$

with some $k \neq 0$, the same for all $f_i^*(x)$ in (32).

By property 2 (21), for each of the random functions $F_1(x), F_2(x), \ldots F_q(x)$, the realizations $f_i(x)$ and $f_i^*(x)$ are equally probable. Therefore, observing $f_Q(x)$ as the combination (30) is as probable as observing $f_Q^*(x)$ as the corresponding combination (31).

Using (32), we can rewrite (31) as

$$f_Q^*(x) = k f_q \left( k f_{q-1} \left( \ldots k f_2 \left( \frac{k f_1 \left( \frac{x}{k} \right)}{k} \right) / k \right) / k \right) =$$

$$= k f_q \left( f_{q-1} \left( \ldots \left( f_1(x/k) \right) \right) \right) = k f_Q(x/k) \qquad (33)$$

Since (33) holds for any realizations and their combinations in (30) - (31), and for any $k$, we can expect that the composite random function (29) inherits the same scale-invariance property (21) as its constituent random functions.

Now consider the correspondence of the composite function (29) to Corollary 3.2 (23). In principle, any continuous real function involving a complex nonlinear dependence on all $n$ input variables could arise as a realization of (29). For example, such a function might already appear as one of the realizations of the first-layer random function $F_1(x)$, while the subsequent realizations $F_2(x), \ldots F_q(x)$ might happen to be linear.

However, since all random functions $F_1(x), F_2(x), \ldots F_q(x)$ share the same characteristics and the realizations are random, it is more plausible to expect combinations where the assembly and increasing complexity of the composite random function (29) proceed layer by layer. Moreover, if, as assumed, the number of outputs $m$ of $F_1(x)$ (the dimensionality after the first layer) is much smaller than the original feature dimension $n$ ($m \ll n$), then we are modeling a situation in which any complex nonlinear dependencies, if present, are determined within a lower-dimensional internal feature space of size $m$, to which we transition in a comparatively simple way using just one layer $F_1(x)$.

Thus we obtain a model consistent with what is commonly expected in practical regression problems. As argued above, in this case Corollary 3.2 (23) may fail.

Such a breakdown of Gaussianity under composition is not unique to our setting. In the literature on Deep Gaussian Processes it is shown that the composition of Gaussian processes is, in general, not a Gaussian process (Damianou & Lawrence, 2013 [6]). Hence, although each layer of the cascade can be interpreted individually as a realization of a Gaussian measure with symmetries (20) - (23), the overall composition can induce a non-Gaussian measure.

In summary, we may hypothesize that if a random function satisfies the symmetry properties of the probability measure with respect to translation, rotation, and scaling (items 1-2, (20) - (21)), but Gaussianity (item 3) is violated in such a way that Cor. 3.1 holds while Cor. 3.2 fails – precisely in situations where one a priori expects the regression solution to lie in a space of lower intrinsic dimension than the original feature space – then a cascade of packages of polyharmonic splines based on (1) - (7) can serve as a suitable modeling framework.

If all intermediate "true" transformed values of the inputs after each package were known, one could directly compute all functions in every layer using (8) - (19). In supervised regression, however, the training set consists only of inputs and their corresponding outputs; the intermediate quantities inside the model between layers are unknown.

Therefore, based on the above reasoning, it is reasonable to expect that chaining packages into a cascade can improve solution quality and open a path to scaling the approach of [1]. We anticipate that developing this mathematical construction and methods for applying it to machine learning problems will make it suitable for "big data."

There exist other approaches to scaling kernel methods. For example, the Nyström method (Williams & Seeger, 2001 [13]) approximates the kernel via a subsample of inducing points, and Fast Multipole Methods (FMM, Beatson & Greengard, 1997 [3]) accelerate computations by hierarchical clustering of the space. These methods, however, typically introduce approximations that may compromise global smoothness of the solution or require heuristic selection of subsets. In contrast, the proposed cascade preserves the exact polyharmonic kernel at every level while providing a natural form of dimensionality reduction – especially beneficial when the feature space is redundant.

The forward pass through the cascade is straightforward. To compute the result for a given set of inputs, first evaluate the output of the first package. Feed this result into the second package, and so on. In other words, one simply evaluates the sequence of packages in the order they are connected in the cascade, using formulas (16) - (19).

We now turn to differentiation. Consider a package of polyharmonic splines used as one component within a larger computational system.

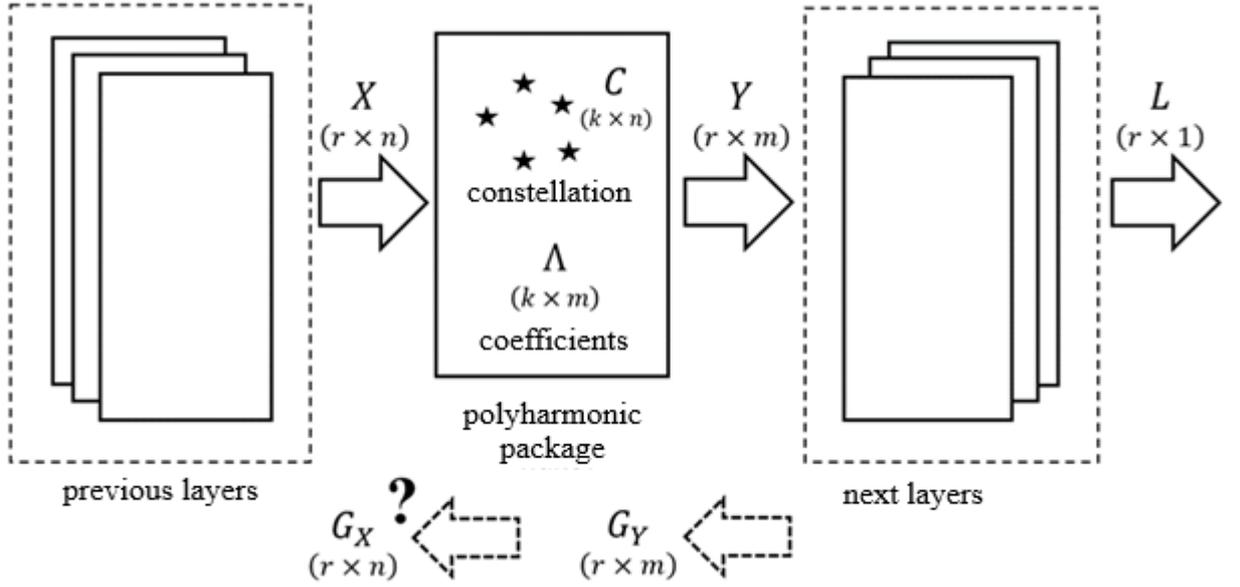

Figure 3.

Let $X$ be the $r \times n$ matrix of elements $x_{ij}$ containing a fragment of data (a batch) passed to the package from previous layers. Each row $x_i = (x_{i1}, x_{i2}, \ldots, x_{in})$ of $X$ is an independently processed vector; the package evaluates $m$ functions on it, forming the row $y_i = (f_1(x_i), f_2(x_i), \ldots, f_m(x_i))$ in the output matrix $Y$. Thus each entry of $Y$ is $y_{it} = f_t(x_i)$.

The rows of $C(k \times n)$ are the constellation points ($k$ points). The matrix $\Lambda$ contains the coefficients used to compute the values $f_t(x_i)$ via (19).

Suppose that somewhere at the system's output we obtain a vector $L = (l_1, l_2, \ldots, l_r)$ of size $r \times 1$ such that each component $l_i$ results from the downstream computation applied to the corresponding input vector $y_i$. That is, each row $y_i$ of $Y$ (independently of the others) produces an output value $l_i$, and the entire matrix $Y$ produces the vector $L$.

Now assume we have somehow obtained the matrix $G_Y$ of size $r \times m$ with entries

$$g_{it}^{(Y)} = \frac{dl_i}{dy_{it}}, \quad i = \overline{1,r} \, ; t = \overline{1,m} \tag{34}$$

In other words, if we regard all subsequent layers as a single $m$-variate function, then $g_{it}^{(Y)}$ are its partial derivatives evaluated at the points $y_i$.

We now wish to compute the matrix $G_X$ with entries

$$g_{it}^{(X)} = \frac{dl_i}{dx_{ij}}, \quad i = \overline{1,r}; j = \overline{1,n} \qquad (35)$$

Having such a procedure allows us to compute derivatives of the overall function represented by the cascade by propagating them backward, layer by layer, from the end to the beginning.

Formally, the procedure for propagating derivatives from layer to layer resembles backpropagation ([2] and [11]). However, in this context it is not used to minimize error, but rather to ensure differentiability of the entire computational chain, which enables integrating the architecture into modern automatic differentiation frameworks.

The elements $x_{ij}$ of the matrix $X$ taken from the i-th row influence only the elements $y_{it}$ from the same i-th row of $Y$, and therefore only the i-th component $l_i$ of the vector $L$.

Since $g_{it}^{(Y)}$ are the partial derivatives of the function that maps $y_{it}$ to $l_i$ at the point $y_i$, the differential of $l_i$ is

$$dl_i = \sum_{t=1}^{m} g_{it}^{(Y)} dy_{it} \qquad (36)$$

From this we obtain $g_{ij}^{(X)}$:

$$g_{ij}^{(X)} = \frac{dl_i}{dx_{ij}} = \sum_{t=1}^{m} g_{it}^{(Y)} \frac{dy_{it}}{dx_{ij}} = \sum_{t=1}^{m} g_{it}^{(Y)} \frac{df_t(x_i)}{dx_{ij}} \qquad (37)$$

Using (9), we can write

$$f_t(x_i) = \sum_{p=1}^{k} \lambda_{pt} k_f(c_p - x_i) \qquad (38)$$

Hence,

$$\frac{df_t(x_i)}{dx_{ij}} = \sum_{p=1}^{k} \lambda_{pt} \frac{dk_f(c_p - x_i)}{dx_{ij}} \qquad (39)$$

The term $k_f(c_p - x_i)$ can be expressed using (12) and (18) as

$$k_f(c_p - x_i) = \frac{1}{2} m_{ip}^{(xc)} \left( \ln\left(m_{ip}^{(xc)}\right) - 2b \right) + c, \tag{40}$$

where $m_{ip}^{(xc)}$ is an entry of the squared-distance matrix (16), which can be written as

$$m_{ip}^{(xc)} = \|c_p - x_i\|^2 = \sum_{u=1}^{n} (c_{pu} - x_{iu})^2 \tag{41}$$

Therefore,

$$\frac{dk_f(c_p - x_i)}{dx_{ij}} = \frac{dk_f(c_p - x_i)}{dm_{ip}^{(xc)}} \cdot \frac{dm_{ip}^{(xc)}}{dx_{ij}} \tag{42}$$

Consider the two factors in (42) separately:

$$\frac{dk_f(c_p - x_i)}{dm_{ip}^{(xc)}} = \frac{1}{2}\left( \ln\left(m_{ip}^{(xc)}\right) - 2b + 1 \right) \tag{43}$$

$$\frac{dm_{ip}^{(xc)}}{dx_{ij}} = \frac{d\left(\sum_{u=1}^{n}(c_{pu} - x_{iu})^2\right)}{dx_{ij}} = 2(x_{ij} - c_{pj}) \tag{44}$$

Substituting (43) and (44) into (39) yields

$$\frac{df_t(x_i)}{dx_{ij}} = \sum_{p=1}^{k} \lambda_{pt}(x_{ij} - c_{pj})\left( \ln\left(m_{ip}^{(xc)}\right) - 2b + 1 \right) \tag{45}$$

Then (37) becomes

$$g_{ij}^{(X)} = \sum_{t=1}^{m} g_{it}^{(Y)} \sum_{p=1}^{k} \lambda_{pt}(x_{ij} - c_{pj})\left( \ln\left(m_{ip}^{(xc)}\right) - 2b + 1 \right) \tag{46}$$

In (46) we can swap the order of summation and split the expression into two terms:

$$g_{ij}^{(X)} = x_{ij} \sum_{p=1}^{k} \left( \ln\left(m_{ip}^{(xc)}\right) - 2b + 1 \right) \sum_{t=1}^{m} g_{it}^{(Y)} \lambda_{pt}$$

$$- \sum_{p=1}^{k} c_{pj} \left( \ln\left(m_{ip}^{(xc)}\right) - 2b + 1 \right) \sum_{t=1}^{m} g_{it}^{(Y)} \lambda_{pt} \tag{47}$$

Introduce matrices $\Theta$ and $\Psi$ of size $r \times k$ (the same size as $M_{xc}$ in (16)), with entries defined by

$$\theta_{ip} = \ln\left(m_{ip}^{(xc)}\right) - 2b + 1, \tag{48}$$

where $m_{ip}^{(xc)}$ are the entries of $M_{xc}$ (16),

and

$$\psi_{ip} = \theta_{ip} \sum_{t=1}^{m} g_{it}^{(Y)} \lambda_{pt} \tag{49}$$

Note. It may happen in (48) that $m_{ip}^{(xc)} = 0$, making the expression undefined. This corresponds to evaluating exactly at a constellation point. In practice, one can substitute a very small positive number for $m_{ip}^{(xc)}$. The precise value in (48) is not critical, because (46) shows it will be multiplied by zero anyway (since $x_{ij} - c_{pj}$ also becomes zero in that case).

Because the sum of products $g_{it}^{(Y)} \lambda_{pt}$ in (49) is simply the matrix product $G_Y \Lambda^T$, we obtain a matrix form for $\Psi$:

$$\Psi = \Theta \circ (G_Y \Lambda^T), \tag{50}$$

where $\circ$ denotes the Hadamard (elementwise) product.

Equation (47) can be written as

$$g_{ij}^{(X)} = x_{ij} \sum_{p=1}^{k} \psi_{ip} - \sum_{p=1}^{k} c_{pj} \psi_{ip} \tag{51}$$

In matrix form this is

$$G_X = X \circ \left((\Psi J_{k,1}) J_{1,n}\right) - \Psi C, \tag{52}$$

where $\circ$ is the Hadamard product,

$J_{k,1}$ is the k×1 vector of ones,

$J_{1,n}$ is the 1×n row vector of ones.

To summarize.

Consider a package of functions (a special form of polyharmonic splines) described by a matrix C containing the constellation points, and by the matrix Λ computed in (10), which holds the coefficients for the functions in the package. The input is a batch of data given by the matrix X.

We can compute the matrix of squared distances $M_{xc}$ between the row vectors of $X$ and the rows of $C$ using (16) - (17), and then form the matrix $K^{(xc)}$ via (18), which contains the values of the correlation function. Multiplying $K^{(xc)}$ by Λ as in (19) yields the output matrix $Y$ of the package.

Incidentally, $K^{(xc)}$ can also be expressed through the matrix Θ whose entries are defined in (48):

$$K^{(xc)} = \frac{1}{2} M_{xc} \circ \left(\Theta - J_{r,k}\right) + c \qquad (53)$$

where $J_{r,k}$ is the $r \times k$ matrix of ones.

If we need partial derivatives at the points specified in $X$ for some larger function that includes the present package as a component, and we are given the partial derivatives at the corresponding rows of $Y$ (the package's outputs), then we can carry out the backpropagation using formulas (48) - (52).

This enables chaining packages into a cascade to form a multi-level computational system. By evaluating the packages in sequence, we obtain the overall system output; similarly, we can differentiate by successively propagating partial derivatives in reverse order from the last package back to the first. Evidently, at the final layer (where the vector $L$ is computed), to initialize the procedure one should take $G_Y$ to be a vector of size $r \times 1$ filled with ones.

From the derived transformations it is clear that all computations reduce to matrix operations, and therefore can be efficiently implemented with parallel computation on GPUs.

Moreover, all arrays are two-dimensional; no three-dimensional or higher-order tensors are required anywhere in the pipeline. This suggests an additional possibility: suppose we have several computational systems with the same structure and with identical constituent packages of splines (the same number of constellation points). The functions computed in these systems – at all stages inside their packages – may be completely different (the matrices $C$ and Λ differ), yet to simulate such a multi-system setup it suffices to move from 2D matrices to 3D tensors. All formulas, for both the forward pass and differentiation, remain the same.

**Conclusion.**

This work proposes a computationally efficient and theoretically motivated architecture based on a cascade of packages of polyharmonic splines. Unlike heuristic deep models, each layer of the cascade retains a rigorous interpretation via random function theory and inherits the kernel derived from symmetry principles in [1]. The presented forward and backward procedures allow seamless integration of this architecture into modern machine learning frameworks. Methods for training the cascade's parameters and an experimental assessment on benchmarks will be presented in subsequent publications.

**Acknowledgments.**

The author thanks the anonymous reviewers for valuable feedback on earlier versions of this work.